\documentclass[final,3p,times,twocolumn]{elsarticle}

\usepackage{subfig}
\usepackage{multirow}
\usepackage{threeparttable}
\usepackage{amsmath}

\journal{Pattern Recognition}

\begin{document}

\begin{frontmatter}



\title{Illumination-aware Faster R-CNN for Robust Multispectral Pedestrian Detection}


\author{Chengyang Li}
\ead{licy\_cs@zju.edu.cn}

\author{Dan Song}
\ead{songdan1992@zju.edu.cn}

\author{Ruofeng Tong\corref{cor1}}
\ead{trf@zju.edu.cn}
\cortext[cor1]{Corresponding author}

\author{Min Tang\corref{}}
\ead{tang\_m@zju.edu.cn}

\address{State Key Lab of CAD\&CG, Zhejiang University, Hangzhou, Zhejiang, China}

\begin{abstract}
Multispectral images of color-thermal pairs have shown more effective than a single color channel for pedestrian detection, especially under challenging illumination conditions. However, there is still a lack of studies on how to fuse the two modalities effectively. In this paper, we deeply compare six different convolutional network fusion architectures and analyse their adaptations, enabling a vanilla architecture to obtain detection performances comparable to the state-of-the-art results. Further, we discover that pedestrian detection confidences from color or thermal images are correlated with illumination conditions. With this in mind, we propose an Illumination-aware Faster R-CNN (IAF R-CNN). Specifically, an Illumination-aware Network is introduced to give an illumination measure of the input image. Then we adaptively merge color and thermal sub-networks via a gate function defined over the illumination value. The experimental results on KAIST Multispectral Pedestrian Benchmark validate the effectiveness of the proposed IAF R-CNN.

\end{abstract}

\begin{keyword}
multispectral pedestrian detection \sep illumination-aware \sep gated fusion


\end{keyword}

\end{frontmatter}


\section{Introduction}
\label{intro}
Pedestrian detection, as a canonical sub-problem of general object detection, has been intensively investigated by the computer vision community \cite{benenson2014ten,nguyen2016human,dollar2012pedestrian}, due to its diversified applications in video surveillance, car safety, image retrieval, robotics, etc. Thanks to the introduction of convolutional neural networks (convnets), especially Faster R-CNN \cite{ren2017faster} and its variants, the past few years have witnessed remarkable improvement in pedestrian detection quality  \cite{hosang2015taking,Zhang_2016_CVPR,li2015scale,zhang2016faster,zhang2017citypersons,cai2016unified,du2017fused}. However, most of current pedestrian detectors are restricted to the benchmarks of color images with good lighting conditions, whereas they probably fail to work under bad illumination conditions, e.g., images captured during nighttime or bad weather days.

Efforts have been invested to deal with bad illumination conditions by considering other types of sensors, such as near infrared cameras, time-of-flight cameras and long-wave infrared (thermal) cameras. Among these sensors, thermal cameras are the most widely used \cite{hwang2015multispectral,lin2015novel,gonzalez2016pedestrian,kim2018pedestrian}, due to their visibility of pedestrian instances in challenging illumination conditions. 
With the appearance of KAIST multispectral pedestrian dataset \cite{hwang2015multispectral}, some approaches designed for color modality are extended to perform multispectral pedestrian detection, including convnets based approaches \cite{wagner2016multispectral,choi2016multi,BMVC2016_73,konig2017fully}.

However, there is a lack of in-depth comparison of different network architectures with respect to necessary adaptations to the pedestrian detection task. It is still unclear what upper limit a vanilla convnet architecture could reach, and in what aspects further improvements are expected. Therefore, in this paper we compare six different convnet fusion architectures which are derived from Faster R-CNN and discuss several potential adaptations. We show that once properly adapted, a vanilla multispectral Faster R-CNN obtains significant improvement from baseline and almost matches the detection performance of the state-of-the-art approach.
 
Intuitively, color and thermal modalities are complementary with each other since they provide different visual cues. Nevertheless, it remains a question how confidently we can rely on each modality.
Our experiments reveal that under good illumination conditions, color and thermal images are complementary with each other; whereas under bad illumination conditions, using thermal images alone is a better choice and fusing with color modality offers no improvement in detection accuracy (details will be given in Section~\ref{arcres}). 
This suggests that the illumination measure can be used as an indicator to facilitate the fusion problem. 

Existing convnets based approaches address the fusion problem of color and thermal modalities mainly via two ways. One is to merge the two streams with equal-weight at score level, regardless of the contributions of the two modalities. This strategy is especially error-prone under bad lighting situations. The other is to fuse the two streams at a specific layer, expecting the network to learn the weighting parameters automatically. However, either image classification or object detection models are tuned to be insensitive to illumination changes, which makes their parameters unsuitable to draw the weighting decision. 
To better handle the fusion problem, a weighting mechanism which takes illumination conditions into account is in demand.

\begin{figure*}[htb]
\centering
  \includegraphics[width=.9\textwidth]{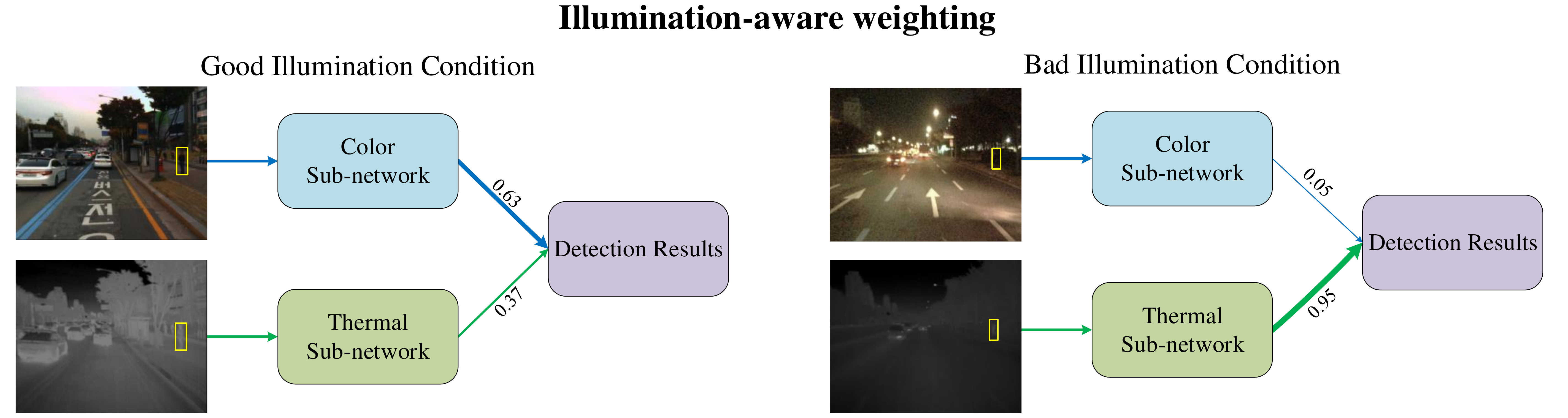}
\caption{Illustration of the illumination-aware weighting mechanism of our IAF R-CNN. A color and a thermal sub-network are responsible for detecting pedestrian instances from color and thermal images respectively. The final detection result is obtained by merging the outputs of the two sub-networks according to the illumination conditions. Left: under good illumination conditions, the weight for color sub-network is higher than that for the thermal sub-network. In this way, color sub-network contributes more than thermal sub-network. Right: under bad illumination conditions, the weight for the thermal sub-network is close to 1. Thus, the final result is dominant by the thermal sub-network. For the sake of conciseness, the region proposal stage is hidden from view in this figure.}
\label{fig:iaw}       
\end{figure*}

Motivated by the above idea, in this paper we develop a novel Illumination-Aware Faster R-CNN (IAF R-CNN) framework. 
We show our weighting mechanism in Fig.~\ref{fig:iaw}.
Given a pair of color-thermal images and pedestrian proposals generated from region proposal network (omitted in the figure for conciseness), color and thermal sub-networks output separate detection confidence scores and bounding box regressions for each proposal.
The final detection result is acquired by merging the outputs of the two sub-networks with an illumination-aware weighing mechanism which consists of two steps.
First, an illumination-aware network is used to offer an illumination measure for the given image. 
Then, the illumination-aware weights for two modalities are predicted by a gate function defined over the illumination measure. It should be noticed that using the proposed IAF R-CNN framework, we can jointly train both multispectral Faster R-CNN and weighting parameters, which makes our method efficient in training.

To sum up, our major contributions are threefold:
(1) We make in-depth comparison of six convnet fusion architectures derived from Faster R-CNN and point out their key adaptations. We reveal that once properly adapted, the performance of a vanilla multispectral Faster R-CNN gains remarkable improvement from baseline and almost matches that of the state-of-the-art approach.
(2) We propose an IAF R-CNN model for multispectral pedestrian detection, which integrates color sub-network, thermal sub-network and a weighting layer into a unified framework. 
(3) We propose an illumination-aware weighting mechanism to lift the contributions of color and thermal sub-networks and boost the final detection performance under both good and bad illumination conditions.
Using the proposed IAF R-CNN, we achieve new state-of-the-art performance on KAIST Multispectral Pedestrian Benchmark.

The remainder of this paper is organized as follows. We first review related work in Section~\ref{related}. Then we compare and analyse six different network fusion architectures and explore their key adaptations in Section~\ref{faster}. Our IAF R-CNN is proposed in Section~\ref{iafrcnn}, followed by extensive experimental results given in Section~\ref{exp}. Finally, we conclude the paper in Section~\ref{conclude}.

\section{Related work}
\label{related}
\noindent \textbf{Convnets for pedestrian detection.}
Driven by the success of convolutional neural networks (convnets) in other classification and detection tasks, the research community has also explored convnets based approaches for pedestrian detection in the past couple of years. 
Most of initial attempts \cite{hosang2015taking,Zhang_2016_CVPR,tian2015deep,tian2015pedestrian,ribeiro2017improving} to apply convnets for pedestrian detection task adopted a two-stage pipeline in an R-CNN \cite{girshick2014rich} style. In these approaches, traditional detectors (usually ICF \cite{BMVC.23.91} and its extensions \cite{dollar2014fast,nam2014local,benenson2013seeking}) were used to generate region proposals, followed by a convnet to re-classify the proposed regions. Other approaches applied convnets in a sliding window manner. ConvNet \cite{sermanet2013pedestrian} used convolutional sparse auto-encoders to initialize the layer parameters, then fine-tuned in a small pedestrian dataset. VeryFast \cite{angelova2015real} built a convnet cascade, in which a tiny convnet is adopted to filter candidates before passing through to a deep convent. F-DNN \cite{du2017fused} used a soft cascade mechanism and fused multiple convnets in the second cascade stage. Although Faster R-CNN \cite{ren2017faster} has become the de-facto standard architecture for general object detection, it under-performed when directly applied to pedestrian detection task, due to low object resolution as well as background confusion \cite{zhang2016faster}. Better performances can be achieved when boosted decision forest was applied on top of convolution feature maps \cite{zhang2016faster,hu2017pushing,cai2015learning}. Other top-performing approaches typically adopted customized convnets derived from Fast/Faster R-CNN architectures \cite{li2015scale,cai2016unified}. Zhang et al. \cite{zhang2017citypersons} revealed that after proper adaptation, a plain Faster R-CNN model can match the state-of-the-art detection performances. 
Recent work \cite{Wang_2018_CVPR,Zhang_2018_CVPR} were built upon Faster R-CNN architecture and focused on occlusion handling.
Wang et al. \cite{Wang_2018_CVPR} proposed a new bounding box regression loss called Repulsion Loss, which particully imporved detection performance in crowd occlusion scenes. 
Zhang et al. \cite{Zhang_2018_CVPR} explored different kinds of attention mechinism, incluing self attention, bounding box based attention and body part based attention, among which body part based attention was the most effecitve one.
Xu et al. \cite{dan2017learning} first trained a RGB-thermal transfer network using multispectral data.
During inference stage only RGB images were used and cross-modal representations were extracted from raw RGB images to perform detection.
Existing convnets based approaches typically focus on color images only. In this paper, we explore Faster R-CNN architecture for multispectral pedestrian detection.

\noindent \textbf{Multispectral pedestrian detection.}
Despite the extensive studies of pedestrian detection task, only a few works have targeted multispectral images. 
Hwang et al. \cite{hwang2015multispectral} extended aggregated channel features (ACF) pedestrian detector \cite{dollar2014fast} and proposed multispectral ACF by augmenting the thermal intensity and HOG feature of the thermal image as additional channel features. 
Wagner et al. \cite{wagner2016multispectral} generated the proposals using multispectral ACF, followed by a late fusion based CNN classifier to re-score them.
Choi et al. \cite{choi2016multi} first generated proposals separate color and thermal stream, and then classified the proposals using support vector regression (SVR) on top of the concatenated convolutional features. 
This pipeline was further extended by Park et al. \cite{park2018unified}, by replacing shallow modules to network components, thus it can be optimized end-to-end.
Liu et al. \cite{BMVC2016_73} introduced Faster R-CNN architecture for multispectral pedestrian detection, and they discussed four different covnet fusion architectures to fuse color and thermal information at different stages.
K{\"o}nig et al. \cite{konig2017fully} applied boosted decision trees to re-score the proposal regions generated by region proposal network. 
Our first contribution is most closely related to \cite{BMVC2016_73}, but we provide better adapted models through making more comprehensive and in-depth investigations. We analyse six fusion architectures while Liu et al. \cite{BMVC2016_73} only investigate four. We examine the impacts of several potential adaptations, enable the detection performances significantly outperform that in \cite{BMVC2016_73}. 

\noindent \textbf{Network fusion problems.}
In this paper we focus on network based multispectral fusion, but we should note that other methods exist than network based ones
(see \cite{jin2017survey,ma2019infrared} for surveys of visible and infrared image fusion).
Although network fusion is less explored for multispectral pedestrian detection \cite{wagner2016multispectral,BMVC2016_73}, it is widely discussed in other vision tasks, such as action recognition \cite{simonyan2014two,karpathy2014large,feichtenhofer2016convolutional}, semantic segmentation \cite{couprie2013indoor,long2015fully,cheng2017locality}, 3D object classification/detection \cite{socher2012convolutional,gupta2014learning}, etc. In \cite{simonyan2014two}, Simonyan et al. proposed two-stream convnets for action recognition, applying two sub-networks to tackle color images and optical flow images separately, which were then combined by averaging the obtained confidence scores. Karpathy et al. \cite{karpathy2014large} discussed several schemes to fuse temporal frames at different speeds for video classification. Cheng et al. \cite{cheng2017locality} introduced a gated fusion layer combining RGB and depth stream according to their varying contributions with respect to different categories and scenes for better semantic segmentation. Targeting scale-variance problem in pedestrian detection, SAF R-CNN \cite{li2015scale} designed a large-scale sub-network and a small-scale sub-network to tackle large instances and small instances respectively, and then fused the two sub-networks by weighting according to the height of proposals. Our IAF R-CNN is similar to SAF R-CNN, as we share a similar divide-and-conquer philosophy. However, our approach is different from theirs in four aspects: (1) SAF R-CNN focuses on a single color channel while our IAF R-CNN targets multispectral images; (2) SAF R-CNN is designed to address the scale-variance problem in single color modality while our IAF R-CNN is proposed to tackle color-thermal fusion problem; (3) SAF R-CNN is built upon Fast R-CNN pipeline \cite{girshick2015fast} while our IAF R-CNN is derived from Faster R-CNN framework \cite{ren2017faster}; (4) our IAF R-CNN adopts a novel weighting mechanism considering the property of specific multispectral pedestrian detection task.

\section{Faster R-CNN for multispectral pedestrian detection}
\label{faster}
\begin{figure*}[htb]
\centering
  \includegraphics[width=.9\textwidth]{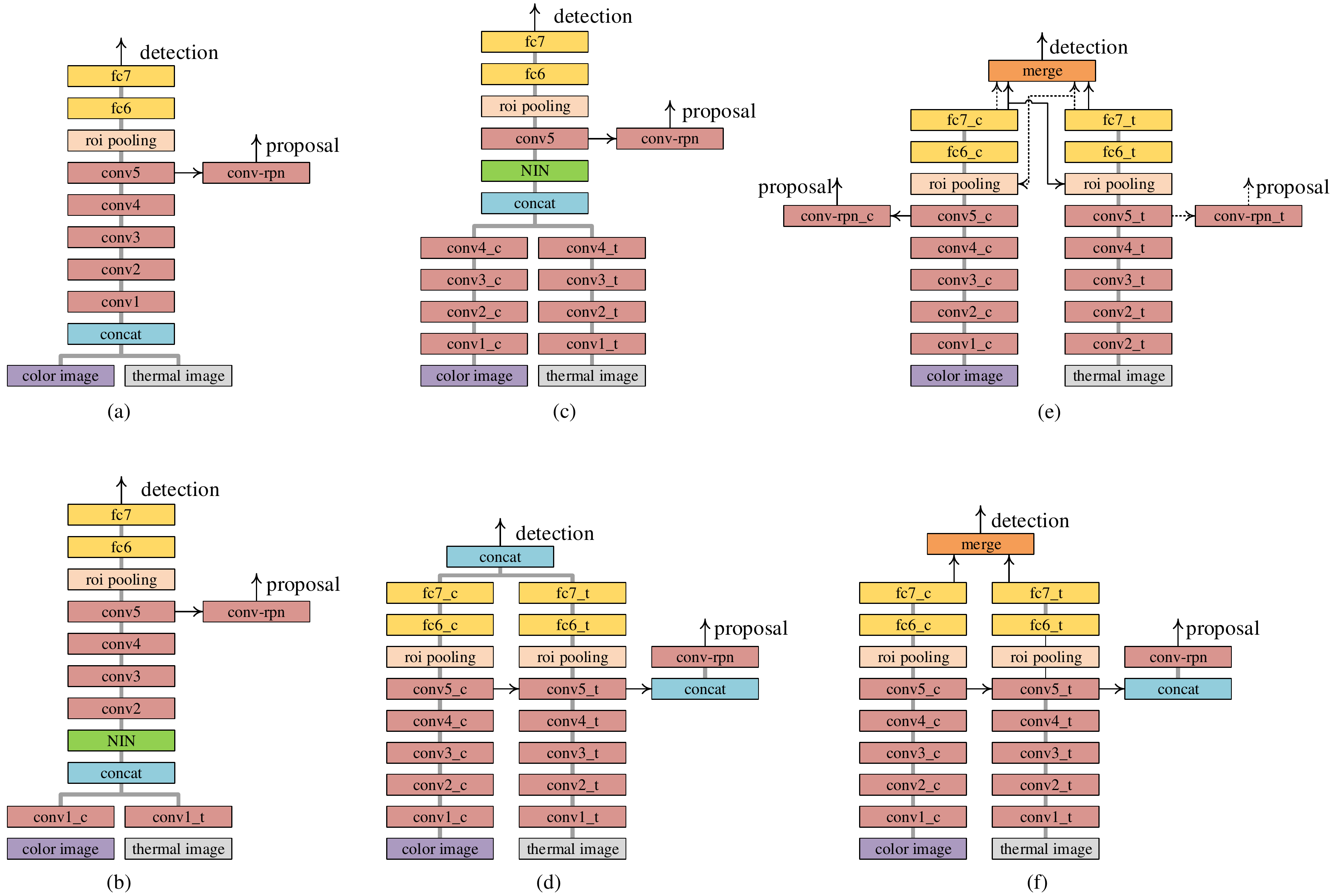}
\caption{We compare six fusion architectures which integrate color and thermal modalities at different stages: (a) Input Fusion, (b) Early Fusion (c) Halfway Fusion (d) Late Fusion (e) Score Fusion I (f) Score Fusion II. For more details, please refer to Section~\ref{arc}.}
\label{fig:arc}       
\end{figure*}

Features at different network stages exhibit different focuses, with finer visual details in lower layers and richer semantic meanings in higher layers. In this section, we make an in-depth comparison of six network fusion architectures derived from Faster R-CNN \cite{ren2017faster}, namely Input Fusion, Early Fusion, Halfway Fusion, Late Fusion, Score Fusion I and Score Fusion II.
As shown in Fig.~\ref{fig:arc}, the six architectures integrate color and thermal modalities at a different stages. 
Four of them have been referred in \cite{BMVC2016_73}, while other two architectures (i.e. Input Fusion and Score Fusion II) are included for more comprehensive study.
These two fusion architectures have been adopted in many vision tasks \cite{wagner2016multispectral,cheng2017locality}, but have not been explored under the framework of Faster R-CNN.
Throughout this paper, we build our networks based on the VGG-16 architecture \cite{simonyan2014very}, and initialize the networks with weights pre-trained on the ImageNet dataset \cite{NIPS2012_4824}. 
Since the models in \cite{BMVC2016_73} is not well adapted for pedestrian detection, here we explore several potential adaptations and point out the key adaptations benefiting the detection performance. The aim of the experiments is not only to seek the upper limits of detection performance of these vanilla architectures, but also to reveal their common limitations and to guide our further improvements.

\subsection{Architectures}
\label{arc}
\noindent \textbf{Input Fusion} simply stacks color and thermal images before feeding them into the network. 
It is the most direct extension of Faster R-CNN from single color modality to color-thermal multi-modality, since only the first convolutional layer needs to be modified due to the increasing of input channels. 

\noindent \textbf{Early Fusion} integrates color and thermal sub-networks right after the first convolutional block, by first concatenating the feature maps from both sub-networks and a subsequent Network-in-Network (NIN) \cite{lin2013network} for dimension reduction. Thus, layers after fusion point can also benefit from pre-trained VGG-16 initialization.

\noindent \textbf{Halfway Fusion} combines color and thermal sub-networks at a later stage, immediately after the fourth convolutional block, via similar feature map concatenation and NIN based dimension reduction.

\noindent \textbf{Late Fusion} is a kind of high-level fusion, which concatenates the last fully-connected layers from color and thermal sub-networks. The feature maps after the last convolutional blocks of the two sub-networks are concatenated, upon which region proposal module is built.

\noindent \textbf{Score fusion I} generates proposals and detections by the two sub-networks separately. The detections are then fed to the other sub-network to re-score the confidence. The final detections are obtained by merging the two-stage detection confidence scores with equal weights of 0.5. Thus, it can be viewed as a cascade design of the two sub-networks.

\noindent \textbf{Score fusion II} is a non-cascade way of fusion at score level. Similar to Late Fusion, human proposals are generated by exploiting feature maps from two sub-networks. 
Then the proposals are taken as input by both sub-networks to generate detection results separately. 
Finally, both detection scores and bounding box regressions from two sub-networks are averaged to obtain the final detections. Compared with Score Fusion I, this execution is more efficient in training and testing.

\subsection{Adaptations}
\label{adapt}
\noindent \textbf{Default setting.}
By default, we mainly follow the original Faster R-CNN \cite{ren2017faster} built upon VGG-16 model but add minor modifications as follows. 
Since we aim to detect standing persons, the anchor ratio of 0.5 is discarded to facilitate the training and testing speed, as is done in \cite{BMVC2016_73}. 
For preparing the training data, we follow \cite{BMVC2016_73} to filter the pedestrian instances by excluding occluded or truncated ones as well as small instances with the height smaller than 50 pixels, resulting in 7,095 training images with a total of 12,790 valid instances. During training, we adopt the image-centric training scheme and use a minibatch consisting 1 image and 120 randomly sampled anchors, with the ratio of positive and negative ones 1:5 \cite{zhang2016faster}. We start training with a learning rate of 0.001, divide it by 10 after 4 epochs, and terminate training after 6 epochs.

\noindent \textbf{Finer feature stride.}
In the default setting, the VGG-16 model has a feature stride of 16 pixels, which is too coarse especially for those small pedestrian instances. To better handle small objects, we remove the last max-pooling layer, providing a finer feature stride of 8 pixels.

\noindent \textbf{Input up-sampling.}
Up-sampling the input images is another strategy to handle coarse feature stride in the default setting. Here, we simply up-sample the input image by a factor of 2.

\noindent \textbf{Include occluded instances.}
The default pedestrian instances for training only include non-occluded ones. However, the reasonable configuration we use for testing consists of both non-occluded and partially-occluded pedestrian instances. As an adaptation, we include partially-occluded instances in training data, which gives 7,601 training images with 14,911 pedestrian instances.

\noindent \textbf{Ignore region handling.}
In the KAIST annotations, there are bounding boxes labelled as \emph{person?} and \emph{people}, which are areas that containing undistinguishable pedestrians and where a human annotator cannot determine if a person is present or not. Additionally, since we only use pedestrian instances with a minimum height of 50 pixels for training, smaller instances are neglected and might be confused as hard negatives. We make sure that these areas are not sampled during training.

\subsection{Multispectral pedestrian detection benchmark}
\label{benchmark}
To the best of our knowledge, KAIST Multispectral Pedestrian Benchmark \cite{hwang2015multispectral} is the only pedestrian dataset that provide large-scale aligned visible and far infrared (FIR, also known as thermal) images with manual annotations. It should be mentioned that the newly published CVC-14 \cite{gonzalez2016pedestrian} is also a multimodal dataset containing visible-FIR image pairs. However, when we visualize the images and the corresponding pedestrian annotations, we find the visible and FIR images are not well aligned. The disparity is particularly noticeable for small pedestrian instances. Thus we use KAIST for analysis in this paper.

The KAIST benchmark consists of 95,328 color-thermal image pairs recorded via a color and a thermal cameras mounted on the rooftop of a car at a equal frame rate of 20 fps. 
The spatial alignment between the two cameras is ensured by a beam splitter and a later camera calibration process.
The manual annotations amount to a total of 103,128 bounding boxes covering 1,182 unique pedestrians. Detection methods are evaluated on a test set consisting of 2,252 images sampled every 20th frame from videos, among which 1,455 images are captured during daytime and the other 797 images during nighttime. The initial procedure for training is to sample every 20th video frame \cite{hwang2015multispectral}. Recent method \cite{BMVC2016_73,konig2017fully} used a finer sampling skip of every 2nd frame to benefit from more training data, which will be adopted in this work. 

As for evaluation, the miss rate (MR) averaged over the range of $[10^{-2}, 10^0]$ false positives per image (FPPI) is taken as the measure of each detection accuracy. The curves of FPPI vs. MR are plotted by the provided evaluation toolbox.

We perform experiments under reasonable configuration \cite{hwang2015multispectral}. The original annotations contain some problematic bounding boxes. Recently, improved annotations were published by Liu et al. \cite{imp_annot}. In this paper, we report the experimental results on both original and improved annotations for more comprehensive comparison. We denote them as $MR^O$ and $MR^I$ for brevity, where O and I stands for ``original annotations" and ``improved annotations" respectively.

\subsection{Results}
\label{arcres}
The step-by-step comparisons of the detection performance are manifested in Table 1. From the table, we have conclusions / findings as follows.

\begin{table*}[htbp]
\newcommand{\tabincell}[2]{\begin{tabular}{@{}#1@{}}#2\end{tabular}}
\tiny
\centering
\caption{\label{comparison}Detection performances (in terms of both $MR^O$ and $MR^I$) of six architectures with different adaptation settings. For each setting, we also report the average performance of all architectures and its improvement from baseline.}
\renewcommand\arraystretch{1.5}
\begin{tabular}{p{0.6cm}p{0.6cm}p{0.6cm}p{0.6cm}p{0.4cm}
p{0.5cm}p{0.5cm}p{0.5cm}p{0.5cm}p{0.5cm}
p{0.5cm}p{0.5cm}p{0.5cm}p{0.5cm}p{0.68cm}}
\hline
\multicolumn{1}{c}{\tabincell{c}{Finer\\ feature\\ stride}} & \multicolumn{1}{c}{\tabincell{c}{Input\\ up-\\scaling}} & \multicolumn{1}{c}{\tabincell{c}{Include\\ occluded\\ instances}} & \multicolumn{1}{c}{\tabincell{c}{Ignore\\ region\\ handling}} & Metric & Color & Thermal & \tabincell{c}{Input\\Fusion} & \tabincell{c}{Early\\Fusion} & \tabincell{c}{Halfway\\Fusion} & \tabincell{c}{Late\\Fusion} & \tabincell{c}{Score\\FusionI} & \tabincell{c}{Score\\FusionII} & Average & \multicolumn{1}{c}{$\Delta$} \\
\hline
\hline
\multicolumn{4}{c}{\multirow{2}{*}{Default setting}} & $MR^O$ &49.27&46.16&41.60&42.68&40.90&38.96&39.02&41.56&40.79&0.00 \\
&&&& $MR^I$ &44.57&31.94&31.04&32.79&29.82&27.36&28.11&30.82&29.99&0.00 \\
\hline
\multicolumn{1}{c}{\multirow{2}{*}{$\surd$}} &&&& $MR^O$ &45.87&43.86&37.70&40.33&34.37&36.08&33.97&36.26&36.45&-4.34 \\
&&&& $MR^I$ &39.42&27.50&26.34&29.38&22.03&24.85&21.49&22.50&24.43&-5.56 \\
\hline
& \multicolumn{1}{c}{\multirow{2}{*}{$\surd$}} &&& $MR^O$ &47.64&42.65&37.24&38.10&32.77&39.13&35.30&39.28&36.97&-3.82 \\
&&&& $MR^I$ &40.98&27.02&23.94&26.79&20.30&24.31&22.67&25.05&23.84&-6.15 \\
\hline
\multicolumn{1}{c}{\multirow{2}{*}{$\surd$}} & \multicolumn{1}{c}{\multirow{2}{*}{$\surd$}} &&& $MR^O$ &51.20&42.95&42.16&41.31&36.86&39.35&35.15&39.27&39.02&-1.77 \\
&&&& $MR^I$ &43.78&26.41&31.49&28.82&20.09&26.75&20.66&24.96&25.46&-4.53 \\
\hline
\multicolumn{1}{c}{\multirow{2}{*}{$\surd$}} && \multicolumn{1}{c}{\multirow{2}{*}{$\surd$}} && $MR^O$ &43.78&41.30&35.69&34.35&30.90&33.70&32.48&32.66&33.30&-7.49 \\
&&&& $MR^I$ &36.51&23.54&24.79&24.08&19.42&20.32&17.80&19.72&21.02&-8.97 \\
\hline
\multicolumn{1}{c}{\multirow{2}{*}{$\surd$}} && \multicolumn{1}{c}{\multirow{2}{*}{$\surd$}} & \multicolumn{1}{c}{\multirow{2}{*}{$\surd$}} & $MR^O$ &43.08&40.70&34.99&35.84&29.99&33.55&32.68&33.05&33.35&-7.44 \\
&&&& $MR^I$ &34.62&22.71&22.47&22.72&17.57&18.89&17.43&18.42&19.58&-10.41 \\
\hline
\end{tabular}
\end{table*}

\begin{figure*}[htb]
\centering
  \includegraphics[width=0.9\textwidth]{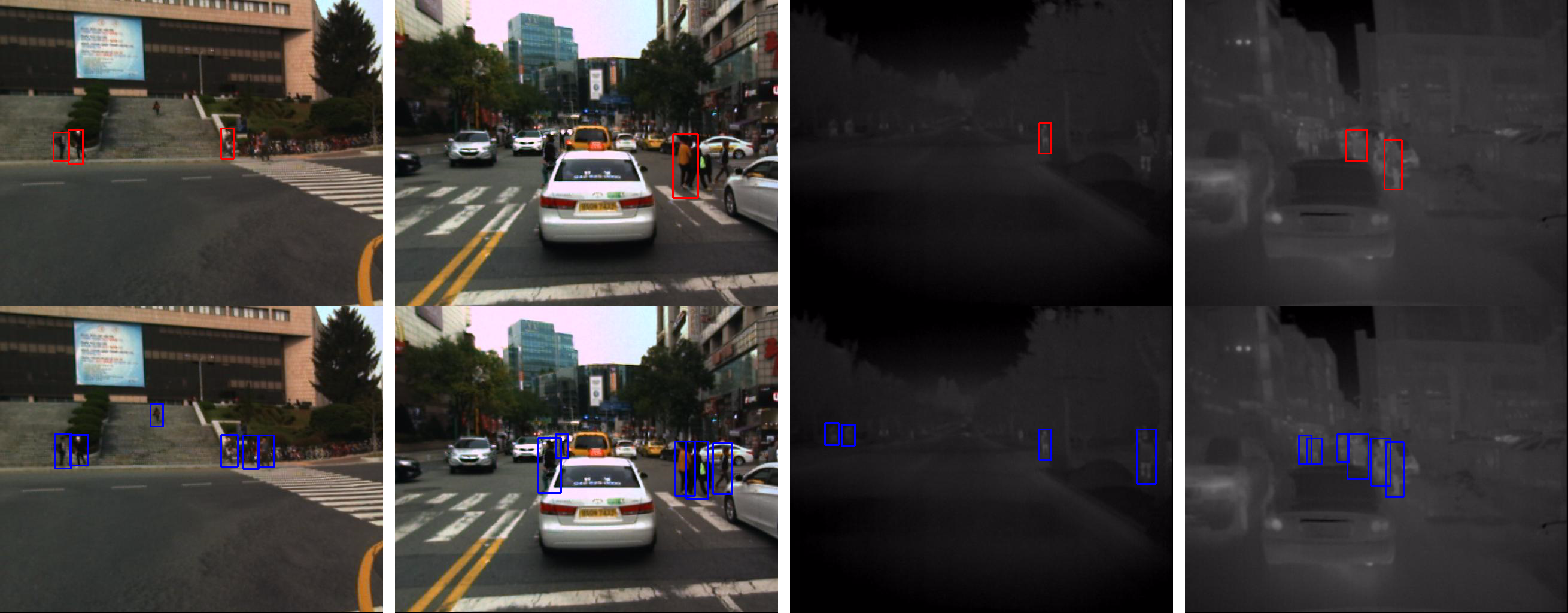}
\caption{Illustration of the original and improved annotations on KAIST test set.
Top: the original annotations. Bottom: our improved annotations.
For the sake of conciseness, we only show color images for daytime scenario and thermal images for nighttime scenario.
The original annotations are relatively coarse and suffer from missing annotation of valid instances.
}
\label{fig:imp}       
\end{figure*}

1. $MR^I$ is more suitable than $MR^O$ to measure the detection performances.
Compared $MR^I$ (using improved annotations) with $MR^O$ (using original annotations), we discover that the value of $MR^I$ is generally lower than that of $MR^O$ by around 10\% to 15\%. 
The overall ranking trends of the two metrics are consistent when $MR^I$ remains high. 
However, when $MR^I$ goes below 25\% or so, the metric of $MR^O$ seems to lose discrimination, as its value oscillates between 30\% and 35\%. 
Close examination of the original test annotations, we find there exist many unlabelled pedestrian instances (see Fig.~\ref{fig:imp} for examples).
With the improvement of the pedestrian detector, those instances not labelled in the original annotations are detected which are then regarded as false negatives when measuring with the original annotations. 
This indicates that $MR^O$ is no longer suitable to measure the detection performance. 
Thus in the rest of this paper, we only measure and report the detection performances in terms of $MR^I$.

2. After proper adaptation, all six fusion architectures gain significant improvements compared with the default setting, with averagely 10.41\% lower in $MR^I$ and 7.44\% lower in $MR^O$. 
Using finer feature stride or input up-sampling alone obtains approximately 6\%, but combining both does not show further improvement. 
Considering the training and testing speeds, we retain finer feature stride and switch off input up-sampling. Adding the rest two adaptations gains further 5\% improvement in $MR^I$.

3. Among the six fusion architectures, Halfway Fusion and Score Fusion I outperform others, achieving 17.57\% and 17.43\% respectively, in terms of $MR^I$. 
The superior performance of Halfway Fusion may benefit from its balance between semantic information and low-level cue, while that of Score Fusion I could attribute to its cascade design. 
Late Fusion and Score Fusion II are just 1\% behind the former two, with 18.89\% and 18.43\% respectively. 
The spatial correspondence is lost when fusing the two sub-networks at fully-connected layer, which could lead to the slightly inferior performance of Late Fusion. 
For Score Fusion II, lack of cascade stage or enough supervision, may explain its inferior performance, compared with Score Fusion I. 
Input Fusion and Early Fusion obtain the worst performance, probably due to the lack of semantic information. 
As of writing, the current best performance on KAIST is RPN+BF \cite{konig2017fully}, with 16.53\% in $MR^I$, and the adapted Halfway Fusion and Score Fusion I just lag by 1\% without using boosting algorithms or other add-ons.

\begin{figure}[htb]
\centering
  \includegraphics[width=.45\textwidth]{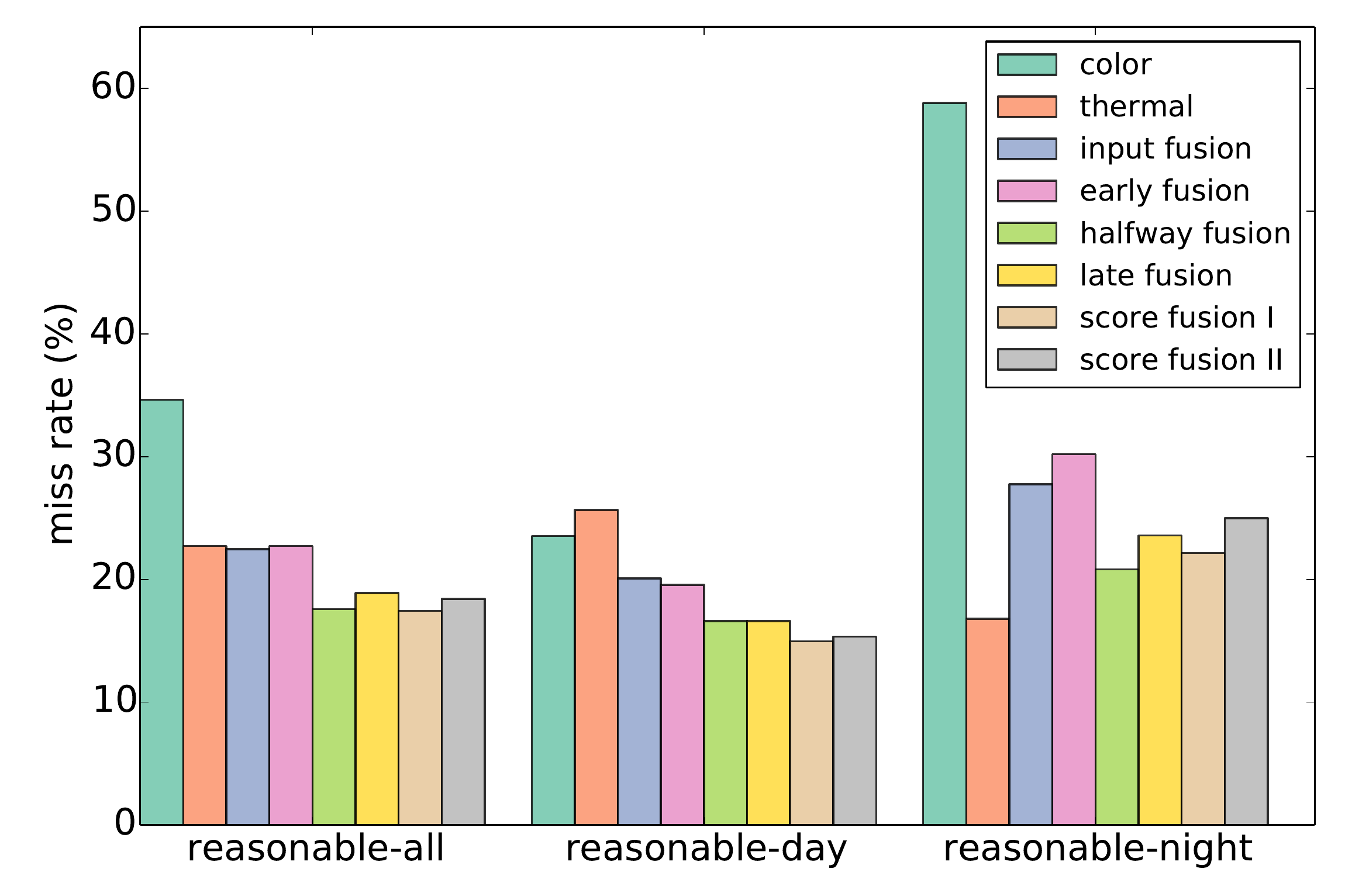}
\caption{Comparison of six fusion architectures as well as color or thermal modality alone under three test configurations, i.e., reasonable-all, reasonable-day and reasonable-night, in terms of $MR^I$.}
\label{fig:arcres}       
\end{figure}

4. Our last finding is about the complementation between color and thermal modalities under different illumination conditions. As shown in Fig.~\ref{fig:arcres}, during daytime, the detection performance of color is slightly better than that of thermal. 
All six fusion architectures obtain better results than using a single modality, indicating color and thermal information are complementary with each other. 
During nighttime, the performance of thermal modality is better than that of color by a large margin, due to the invisibility of visual-optical spectrum at night. 
Nevertheless, it is surprising that all six architectures fail to surpass the result of thermal modality, suggesting color images actually cause confusion rather than offer help for pedestrian detection under bad illumination conditions.

\section{Illumination-aware Faster R-CNN}
\label{iafrcnn}

\subsection{Overall architecture}
\label{overall}
\begin{figure*}[htb]
\centering
  \includegraphics[width=.9\textwidth]{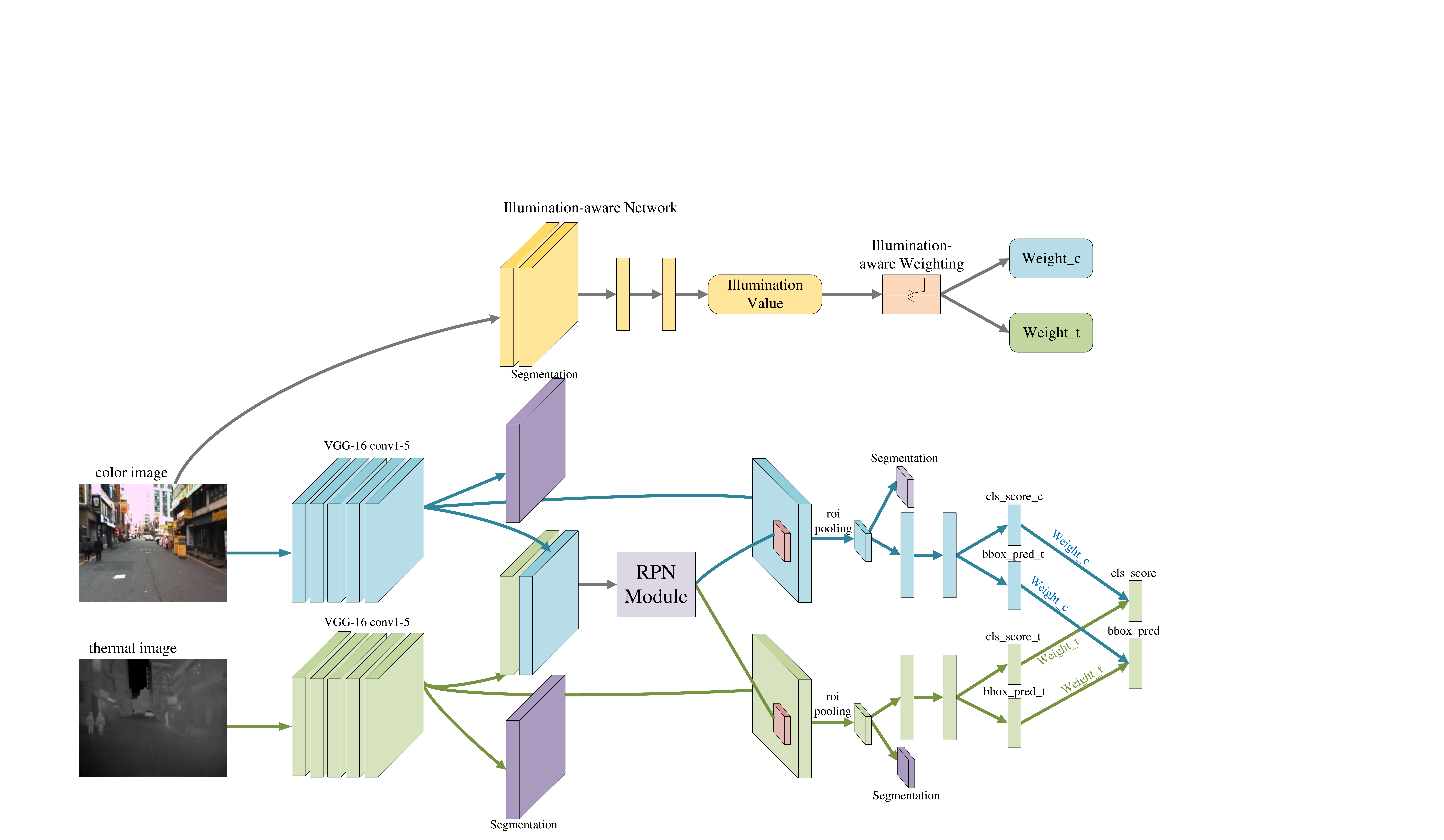}
\caption{The architecture of the proposed IAF R-CNN. Two sub-networks takes color image and thermal image respectively as input and generate separate detections in terms of classification confidence scores and bounding box coordinates. Meanwhile, in a side branch, an illumination-aware network is used to estimate the illumination value from the given color image, followed by an illumination-aware weighting layer to compute the fusion weights for the two modalities via a gate function defined over the estimated illumination value. The final detection results are obtained by weighting the results of the two sub-networks using the computed fusion weights. Purple boxes denote segmentation layers, which are only used during training stage. Best viewed in color.}
\label{fig:overview}       
\end{figure*}

Fig.~\ref{fig:overview} illustrates the overall architecture of the proposed Illumination-aware Faster R-CNN (IAF R-CNN), which is developed based on the Faster R-CNN detection framework \cite{ren2017faster} and our experimental findings discussed in Section~\ref{faster}. IAF R-CNN is composed of three parts: the trunk multispectral Faster R-CNN, the side illumination estimation module, and the final gated fusion layer. A multispectral Faster R-CNN is adopted to generate separate detections from the color image and the thermal image respectively. The illumination estimation module is designed to give an illumination condition measure of the given image. 
Finally, to enable accurate and robust detection, a gated fusion layer is introduced to fuse the color and thermal detection results, which takes the estimated illumination measure into account.

We adopt the architecture of Score Fusion II for the truck multispectral Faster R-CNN model, but remove the original average weighting layer, enabling the output of this stage to be separate detections from the two modalities, in terms of classification confidence scores and bounding box coordinates. 
We choose this fusion type with following two reasons. Compared with fusion at convolutional level or fully-connected level, fusion at score level is more semantic and explicit which can be better weighted. Compared with Score Fusion I, Score Fusion II removes the additional cascade stage, thus is more concise and straightforward. 
To improve the shortcomings of Score Fusion II, we replace the weighting scheme and modify the optimization strategy. 
Besides, we use pedestrian masks as additional supervision following Brazil et al. \cite{brazil2017illuminating}, since they demonstrated its benefits in color image based pedestrian detection. For implementation, the segmentation module is simply a single $1\times1$ convolutional layer.
We consider three different ways to measure illumination conditions given input images, two in a traditional style and one in a network fashion. We find that predicting illumination conditions by Illumination-aware Network (IAN) is the most effective.
IAN consists of a chains of convolutional, fully-connected and max pooling layers, taking the color image as input and providing an illumination condition measure. 
Towards the gated fusion layer, we use a gate function defined over the illumination measure to compute the fusion weights for the two modalities, which will be used for weighting the detection results from the two modalities to obtain the final results. 
More details of the proposed IAF R-CNN will be described in the following subsections.

\subsection{Illumination Estimation}
\label{ian}
Given an image pair, we estimate the illumination conditions from the color image because the thermal image is less sensitive to illumination changes.
Formally, illumination estimation can be defined as the mapping $I \rightarrow iv$, where $I$ denotes an input image and $iv \in [0,1]$ represents an illumination value. 
It is a non-trival task since illumination condition is an ambiguous concept and the ground-truth illumination labels are not available in the pedestrian dataset.
We consider three different illumination measures in our experiment. 

\textbf{Key \& Range.}
According to Kopf et al. \cite{kopf2007capturing}, the luminance characteristics of an image can be measured by its Key (average luminance) and Range.
Specifically, we determine the key as the average pixel value in an image, while the range is the difference between the $90^{th}$ and $10^{th}$ pixel value percentiles.
Finally the Key and the Range are normalized to the interval $[0,1]$.

\begin{figure}[htb]
\centering
  \includegraphics[width=.45\textwidth]{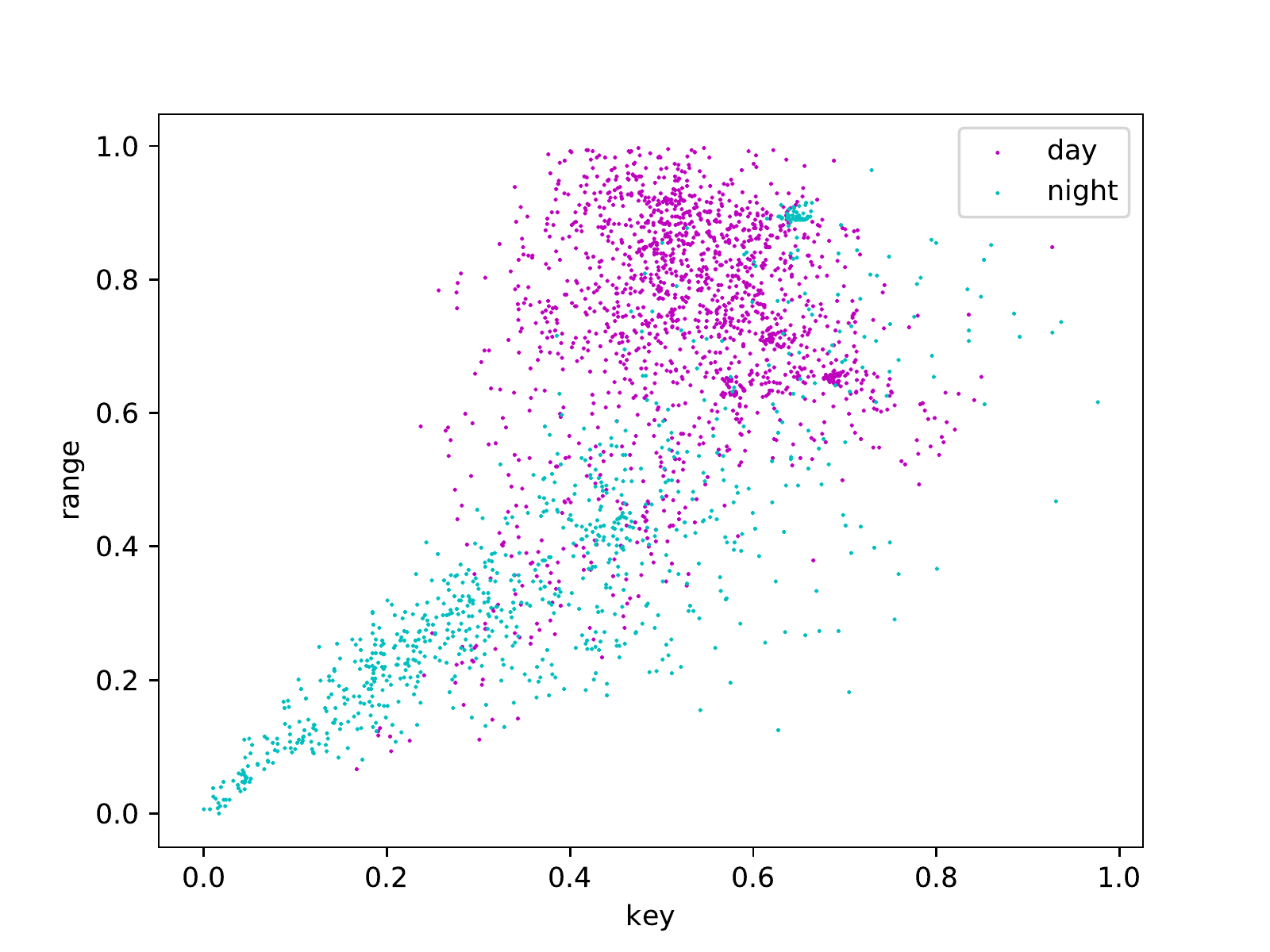}
\caption{Distribution of key and range on KAIST train set, sampled every 20th frame.}
\label{fig:keyrange}       
\end{figure}

The distribution of the Key and the Range in the KAIST train set is illustrated in Fig.~\ref{fig:keyrange}. We can observed that nighttime images generally have relative smaller values than daytime images for both the Key and the Range, but there exist certain overlaps between daytime and nighttime images with regard to these two measures. 

\textbf{IAN.}
We also consider introducing a network, denoted IAN, to estimate the illumination conditions. 
Since there is no ground-truth labels in the dataset, we use the coarse day/night labels instead to train IAN. 

The input color image is resized to $56 \times 56$ pixels to facilitate training and testing efficiency. IAN consists of two convolutional layers with $3 \times 3$ filters, each of which followed by a ReLU layer and a $2 \times 2$ max pooling layer, and two subsequent fully-connected layers with 256 and 2 neurons respectively. A dropout layer with a ratio of 0.5 is inserted after the first fully-connected layer to alleviate over-fitting. The network is trained by minimizing the softmax loss between the prediction and the label, and the softmax score of day category is used as the output illumination value.

It should be noted that rather than reusing the computed features in the trunk multispectral Faster R-CNN, we estimate the illumination value from the color image directly for two reasons. First, the trunk network is pre-trained on image classification task and then fine-tuned on object detection task, however, models in both tasks are adapted to be invariant to illumination changes. Another reason is that we adopts the ``image-centric" sampling strategy in training, while learning the illumination estimation requires large mini-batch to ensure convergence. 

We experimentally determine that IAN is the most effective method for illumination estimation (see Section~\ref{ablation} for details), which will be adopted in our final pipeline.

\subsection{Gated fusion}
\label{gated}
The gated fusion layer is introduced to effectively combine color and thermal for pedestrian detection. 
An illumination-aware weighting mechanism is designed to generate fusion weights for color and thermal modalities according to the illumination conditions. 
As discussed in Section~\ref{arcres}, the fusion weights for color and thermal modalities should satisfy the following constraints. 
Under good illumination conditions, the weight for color sub-network should be high while the weight for thermal sub-network should not be too small, so that the final detection results would benefit from both modalities. 
Conversely, under bad illumination conditions, the weight for the thermal sub-network is supposed to be dominant while that for the color sub-network is supposed to be insignificant, because color images provide more interference than help. 
With these observations in mind, we carefully design a gate function defined over the estimated illumination value $iv \in [0,1]$ as follows. 
\begin{equation}
\label{eq:gatefunction}
w = \frac{iv}{1+\alpha exp(-\frac{iv-0.5}{\beta})}
\end{equation}
where $\alpha$ and $\beta$ are two learnable parameters. We term $w^{color} = w$ and $w^{thermal} = 1-w$ as the weights for fusing the two modalities, where $w^{color}$ and $w^{thermal}$ indicate how confidently we can rely on color and thermal respectively to predict the occurrence of pedestrian instances in the given image. 

Recall that each sub-network generates two outputs: confidence score $s = (s_0,\ldots,s_K)$ over K+1 categories and bounding-box regression offsets $t = (t_1,\ldots,t_K)$ for each of K object categories. Thus, given $s^{color}$ and $t^{color}$ from color sub-network and $s^{thermal}$ and $t^{thermal}$ from thermal sub-network, we obtain the final detection results as
\begin{equation}
\label{eq:swf}
s^{final} = w^{color} \times s^{color} + w^{thermal} \times s^{thermal}
\end{equation}
\begin{equation}
\label{eq:twf}
t^{final} = w^{color} \times t^{color} + w^{thermal} \times t^{thermal}
\end{equation}

\subsection{Optimization}
\label{optimization}
The training procedure of IAF R-CNN consists of two main phases.
In the first phase, we only train the trunk Faster R-CNN by minimizing the following joint loss function with seven terms: 
\begin{equation}
\begin{split}
\label{eq:loss}
\mathcal{L} = &\lambda_1\mathcal{L}_{rpn}
+\lambda_2\mathcal{L}_{dn}^{color}+\lambda_3\mathcal{L}_{dn}^{thermal}\\
+&\lambda_4\mathcal{L}_{seg}^{color}+\lambda_5\mathcal{L}_{seg}^{thermal}
+\lambda_6\mathcal{L}_{seg_{roi}}^{color}+\lambda_7\mathcal{L}_{seg_{roi}}^{thermal}
\end{split}
\end{equation}
where $\mathcal{L}_{rpn}$ is the proposal loss, $\mathcal{L}_{dn}^{color}$ and $\mathcal{L}_{dn}^{thermal}$ are the detection losses of color and thermal sub-networks respectively. The formulation of proposal loss and detection loss remain the same as Faster R-CNN \cite{ren2017faster}.

Following \cite{brazil2017illuminating}, we also introduce two kinds of person segmentation loss in the joint loss function.
$\mathcal{L}_{seg}^{color}$ and $\mathcal{L}_{seg}^{thermal}$ are the image-level per-pixel loss.
Let $G_{x,y}$, $P_{x,y}$ respectively be the ground-truth and predicted segmentation masks, the image-level per-pixel loss is defined as:
\begin{equation}
\mathcal{L}_{seg} = \frac{1}{H \times W}\sum_{(x,y)}l(G_{x,y},P_{x,y})
\end{equation}
where H and W are the size of the feature map and $l$ is the cross-entropy loss function.
$\mathcal{L}_{seg_{roi}}^{color}$ and $\mathcal{L}_{seg_{roi}}^{thermal}$ are the roi-level per-pixel loss.
Let $G_{x,y,c}$, $P_{x,y,c}$ respectively represent the ground-truth and predicted segmentation masks of the $c_{th}$ roi, the roi-level per-pixel loss can be computed as:
\begin{equation}
\mathcal{L}_{seg_{roi}} = \frac{1}{H \times W \times C}\sum_{(x,y,c)}l(G_{x,y,c},P_{x,y,c})
\end{equation}
where C is the number of rois and other notations remain the same as $\mathcal{L}_{seg}$.
In our experiments, we set all $\lambda_i=1$.

In the second phase, we optimize the weighting parameter in the gated function by minimizing the loss function $\mathcal{L} = \mathcal{L}_{dn}^{final}$ where $\mathcal{L}_{dn}^{final}$ is the detection loss defined over the final detections. In this phase, we only propagate back to the gated fusion layer, since full back-propagation does not make further improvement.

\section{Experiments}
\label{exp}

\subsection{Implementation details}
\label{implementation}
Our framework is implemented under Tensorflow \cite{abadi2016tensorflow}. 
For training IAN, we use the color images from the training set of KAIST dataset. The network is initialized with random Gaussian distribution and is trained by Adam solver for 2 epochs with a learning rate of 0.0001 and a batch size of 64. No data augmentation is used during the training. 
For optimizing the parameters in the trunk multispectral Faster R-CNN we use exactly the identical training schedule and hyper-parameters as we described in Section~\ref{faster} for the sake of comparison purpose. For optimizing the parameters in the gated function, we start training with a learning rate of 0.01, divide it by 10 after 2 epochs, and terminate training after 3 epochs.
We set the initial values of parameter $\alpha$ and $\beta$ in Eq.~\ref{eq:gatefunction} to 0.1 and 1 respectively.

\subsection{Comparison with state-of-the-arts}
\label{comp}

\begin{figure*}[thb]
\centering
\subfloat[Reasonable all]{%
  \includegraphics[width=.33\textwidth]{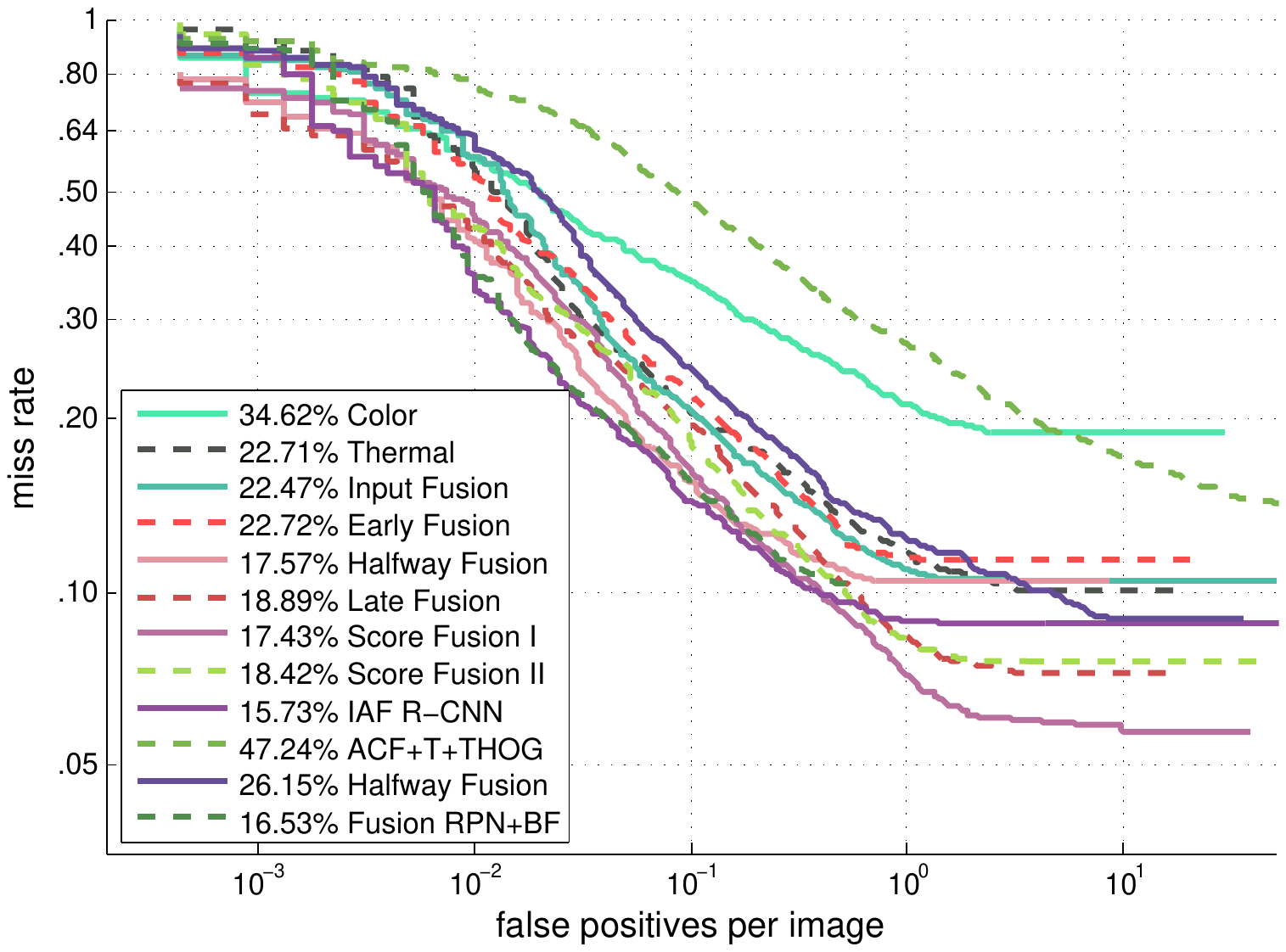}}\hfill
\subfloat[Reasonable day]{%
  \includegraphics[width=.33\textwidth]{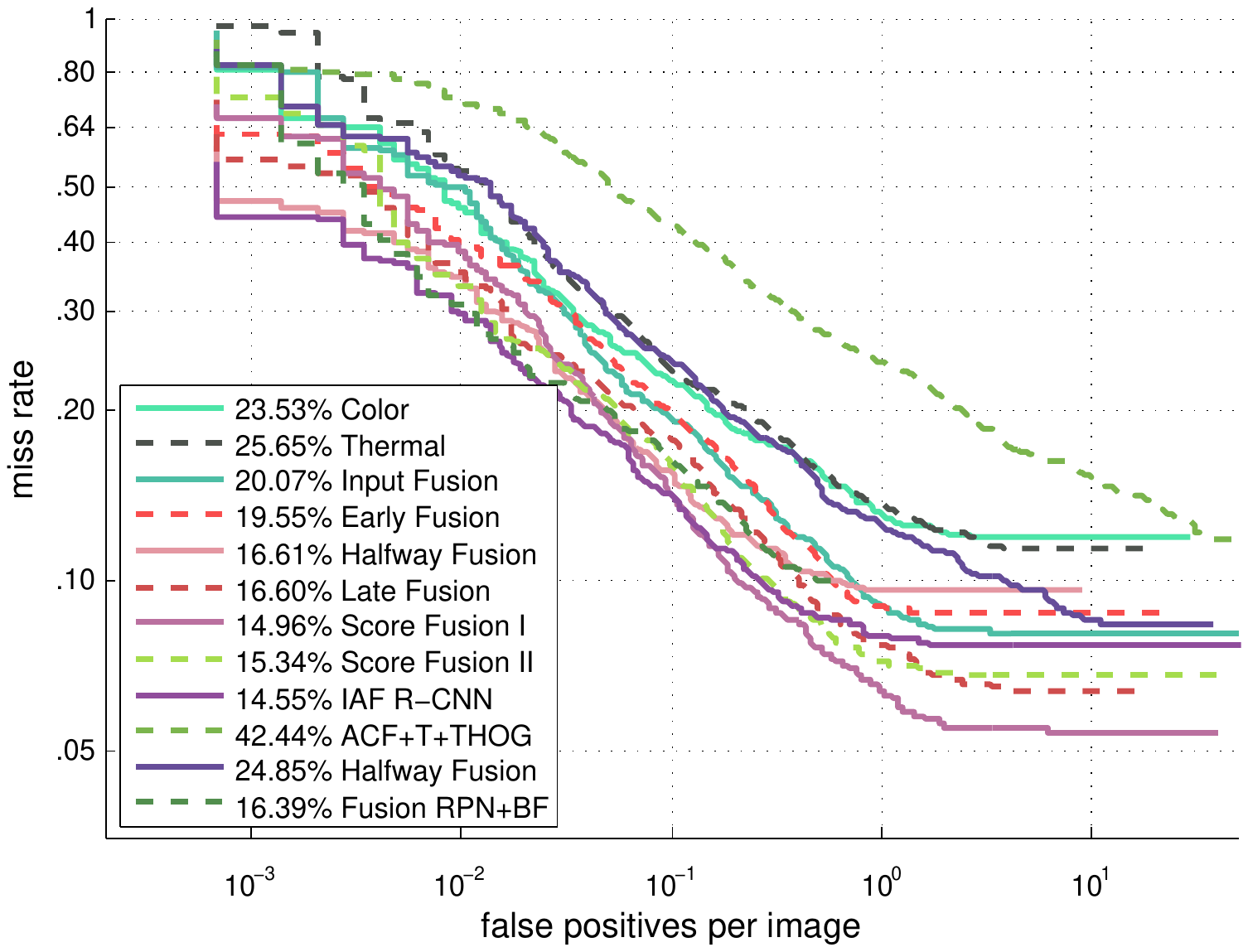}}\hfill
\subfloat[Reasonable night]{%
  \includegraphics[width=.33\textwidth]{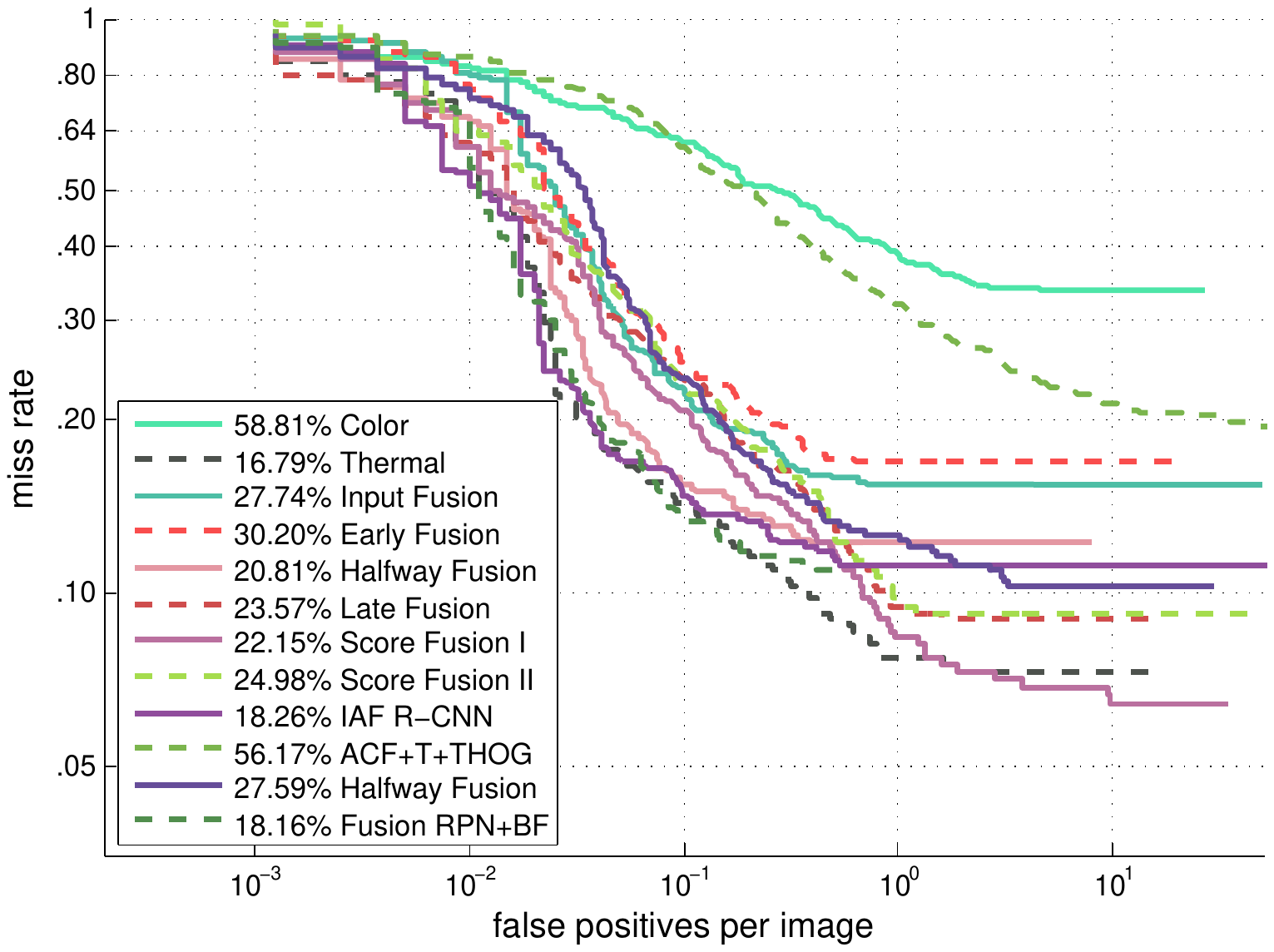}}
\caption{Comparison of detection results ($MR^I$) reported on the test set of KAIST dataset.}
\label{fig:compall}
\end{figure*}

We train our model on the KAIST training set and evaluate it on the KAIST testing set. 
We compare the proposed approach with other published approaches, in terms of $MR^I$ under reasonable configuration \cite{hwang2015multispectral}. 
ROC curves are presented in Fig.~\ref{fig:compall}, where we compare our approach with \cite{hwang2015multispectral,BMVC2016_73,konig2017fully} as well as the architectures we discussed in Section~\ref{faster}. The authors of \cite{hwang2015multispectral,BMVC2016_73,konig2017fully} provide their codes or detections, so we re-evaluate and report their detection performances on the improved test annotations using the toolbox provided by KAIST dataset.
It can be observed that IAF R-CNN outperforms all these methods and achieves the lowest $MR^I$ of 15.73\%. 
From the figure we can observe that during daytime our approach obtains the best performance, while during nighttime using thermal modality alone provide the lowest log-average miss rate and our approach as well as RPN+BF \cite{konig2017fully} have a similar performance of around 18.2\% just second to that of thermal.

Table~\ref{table:time} illustrates the computational cost of our method
compared to the state-of-the-art methods.
It can be observed that the proposed IAF R-CNN is also time-efficient during inference stage, with only 0.21s/image.

\begin{table*}[h]
\centering
\scriptsize
\begin{tabular}{cccccc}
\hline
Methods & Choi et al.\cite{choi2016multi} & Park et al. \cite{park2018unified} & Halfway Fusion \citep{BMVC2016_73} & Fusion RPN+BF \cite{konig2017fully} & IAF R-CNN \\
\hline
Time (s.) & 2.73 & 0.58 & 0.43 & 0.80 & 0.21 \\
\hline
\end{tabular}
\caption{Comparison of computation time using a NVIDIA GeForce GTX TITAN X GPU.}
\label{table:time}
\end{table*}

\subsection{Ablation studies}
\label{ablation}

\subsubsection{Illumination-aware weighting}
\label{ab-iaw}

\begin{figure}[htb]
\centering
  \includegraphics[width=.4\textwidth]{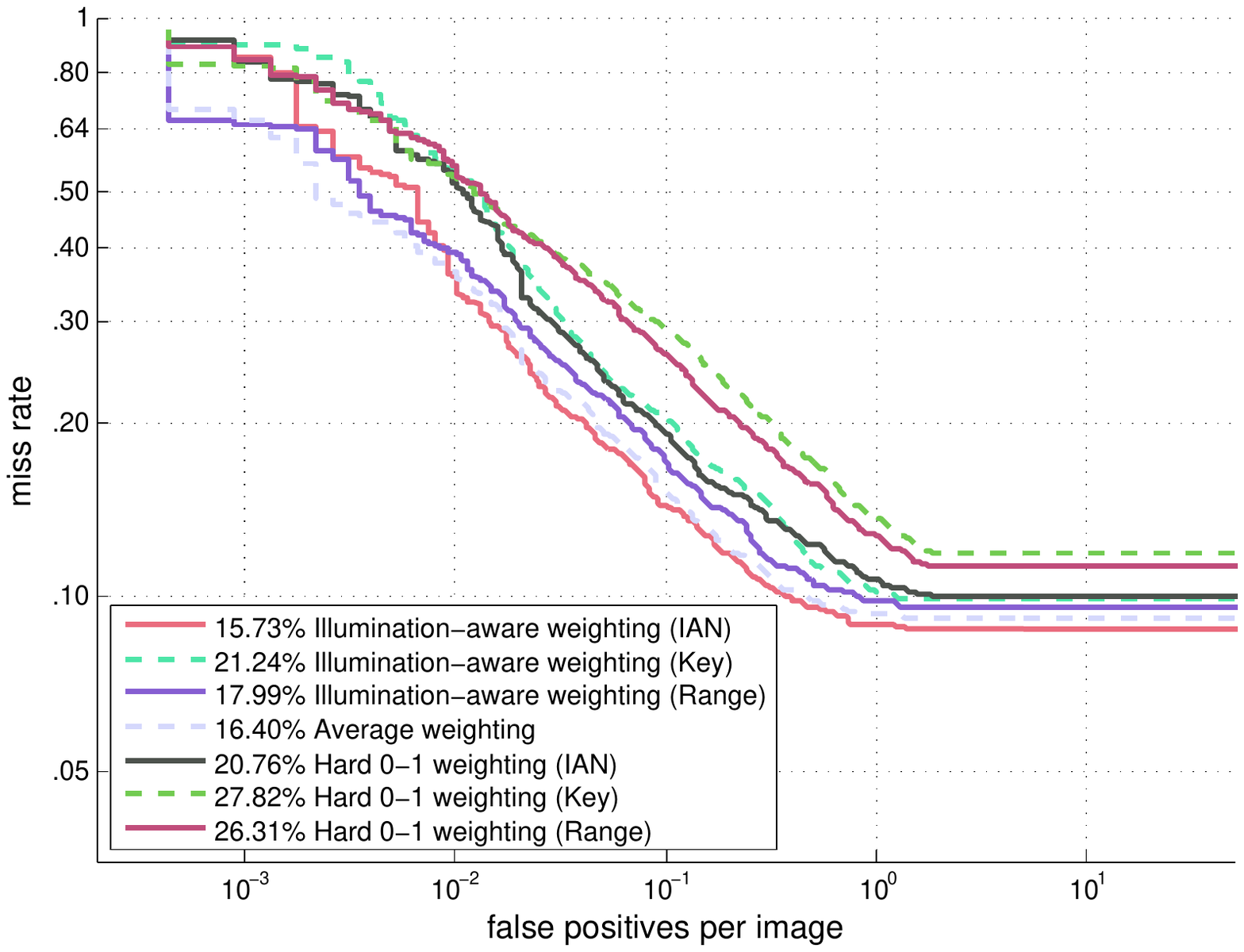}
\caption{Comparison of detection performances (reasonable-all, $MR^I$) with different weighting mechanisms: average weighting, hard 0-1 weighting and illumination-aware weighting as well as illumination estimation methods Key, Range and IAN.}
\label{fig:weighting}       
\end{figure}

To demonstrate the effectiveness of ``illumination-aware weighting", we compare it with other weighting mechanisms, namely ``average weighting" and ``hard 0-1 weighting". For average weighting, we merge the detections from both sub-networks with equal weights of 0.5, while for hard 0-1 weighting, we directly adopt the detections from color sub-network when the illumination value is larger than 0.5, otherwise the opposite. We also test the three illumination estimation methods, i.e, Key, Range and IAN.

Fig.~\ref{fig:weighting} illustrates the comparison of the three weighting mechanisms as well as three illumination estimation methods. 
It can be observed that the results using Key or Range as illumination measure under-perform that using IAN. We believe the reason is that outdoor scenes contain complex and challenging backgrounds, which can not be straightforward handled by pixel value statistics. Although the training procedure of IAN only makes use of day/night labels, IAN is able to learn the distinctive characteristic between daytime and nighttime. If provided more rich illumination labels, the detection performance can be further boosted.
Using IAN as the illumination estimator, the result of the proposed illumination-aware weighting outperforms that of average weighting by 0.67\% and outperforms that of hard 0-1 weighting by 5.03\%, in terms of $MR^I$. It indicates that the proposed illumination-aware weighting mechanism can effectively merge the detections from the two sub-networks according to the illumination conditions and is robust to the illumination changes of the given images.

\subsubsection{Visualization of Illumination-aware Fusion}
\label{visfusion}
\begin{figure*}[thb]
\centering
\subfloat[Examples]{%
  \includegraphics[width=.75\textwidth]{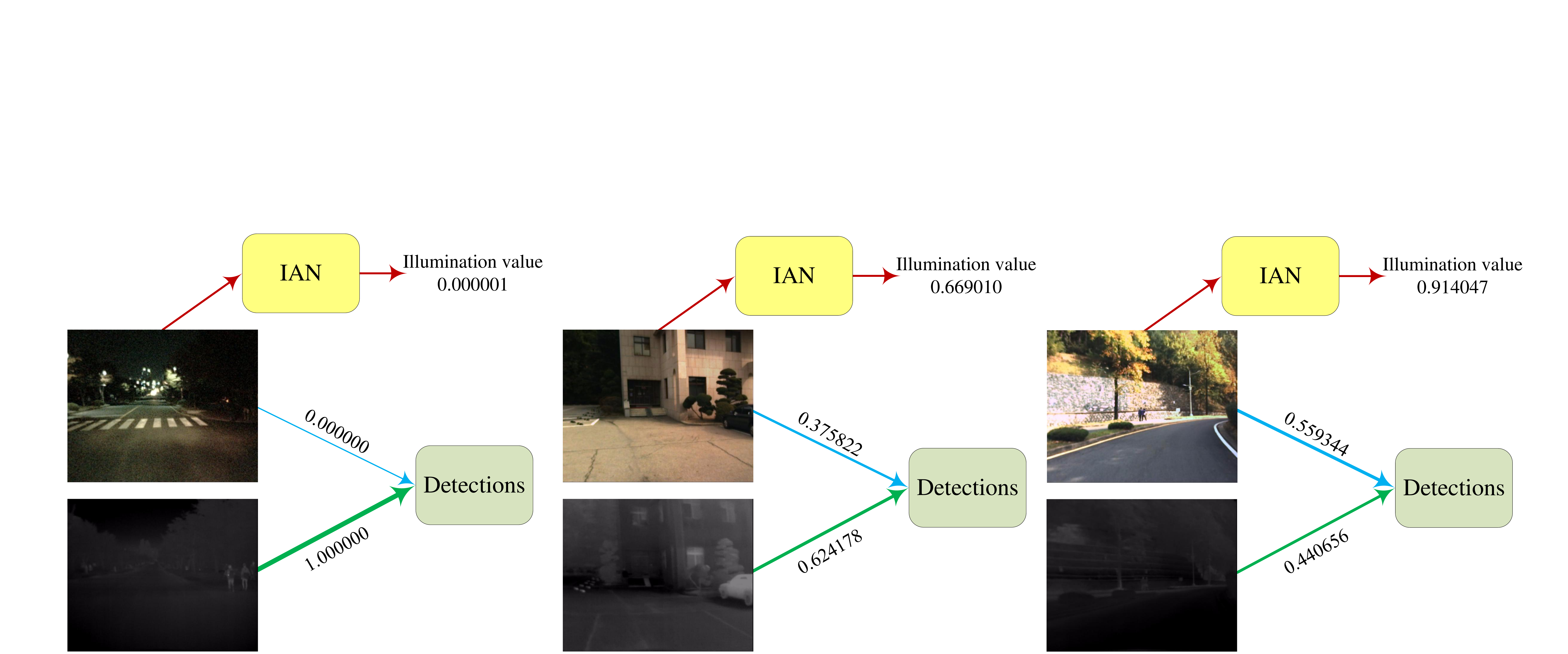}}\hfill
\subfloat[Gated function]{%
  \includegraphics[width=.25\textwidth]{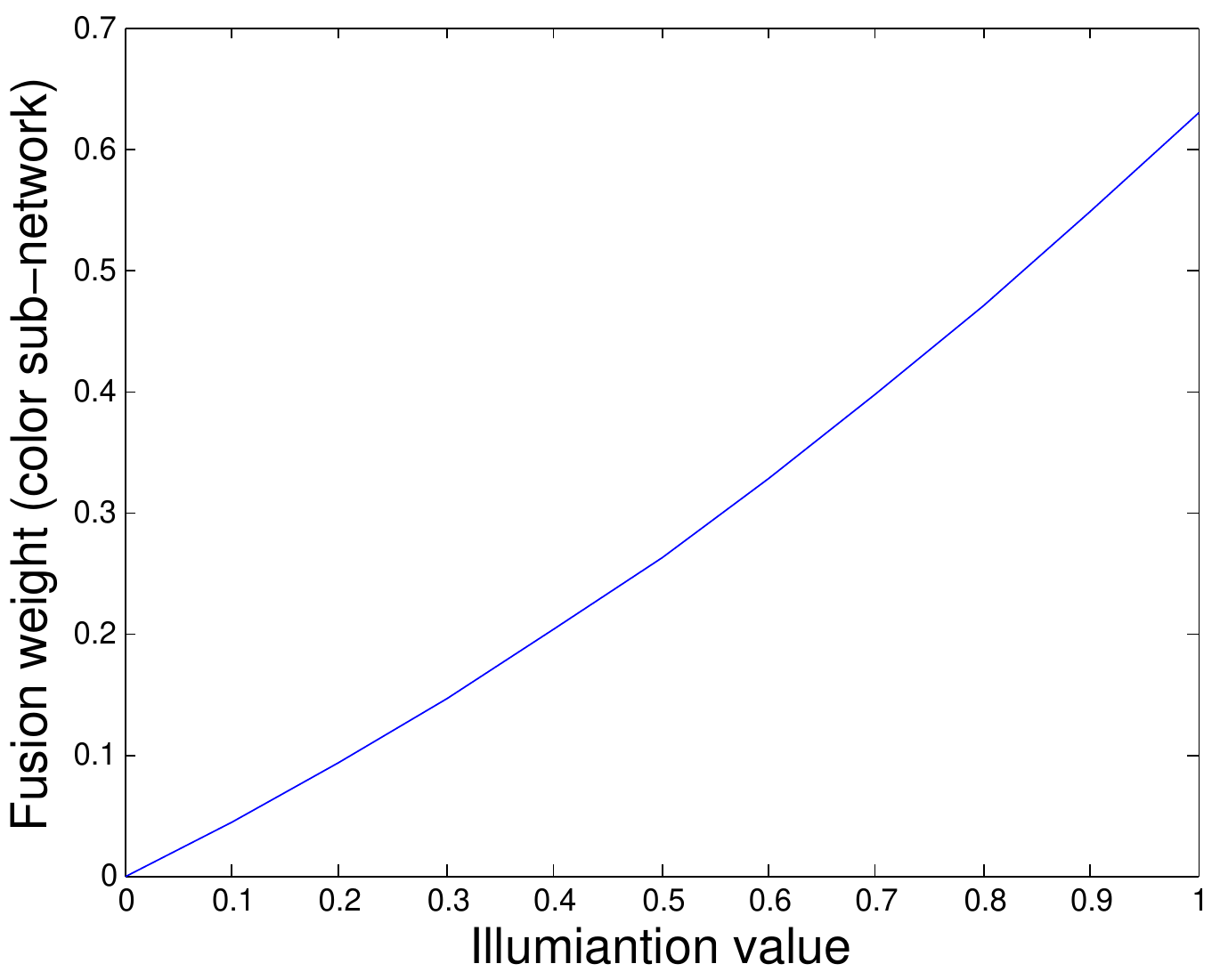}}
\caption{Illustration of illumination-aware examples and gated function.}
\label{fig:visweight}
\end{figure*}

Some examples with regard to the illumination-aware weighting mechanism is illustrated in Fig.~\ref{fig:visweight} (a) and the automatically optimized gated function is plotted in Fig.~\ref{fig:visweight} (b).
It can be observed that the proposed illumination-aware weighting mechanism can adaptively choose weights for the color and thermal network fusion, according to the estimated illumination value generated by the IAN.

\subsubsection{Visualization of detection results}
\label{vis}

\begin{figure*}[h]
\centering
  \includegraphics[width=.95\textwidth]{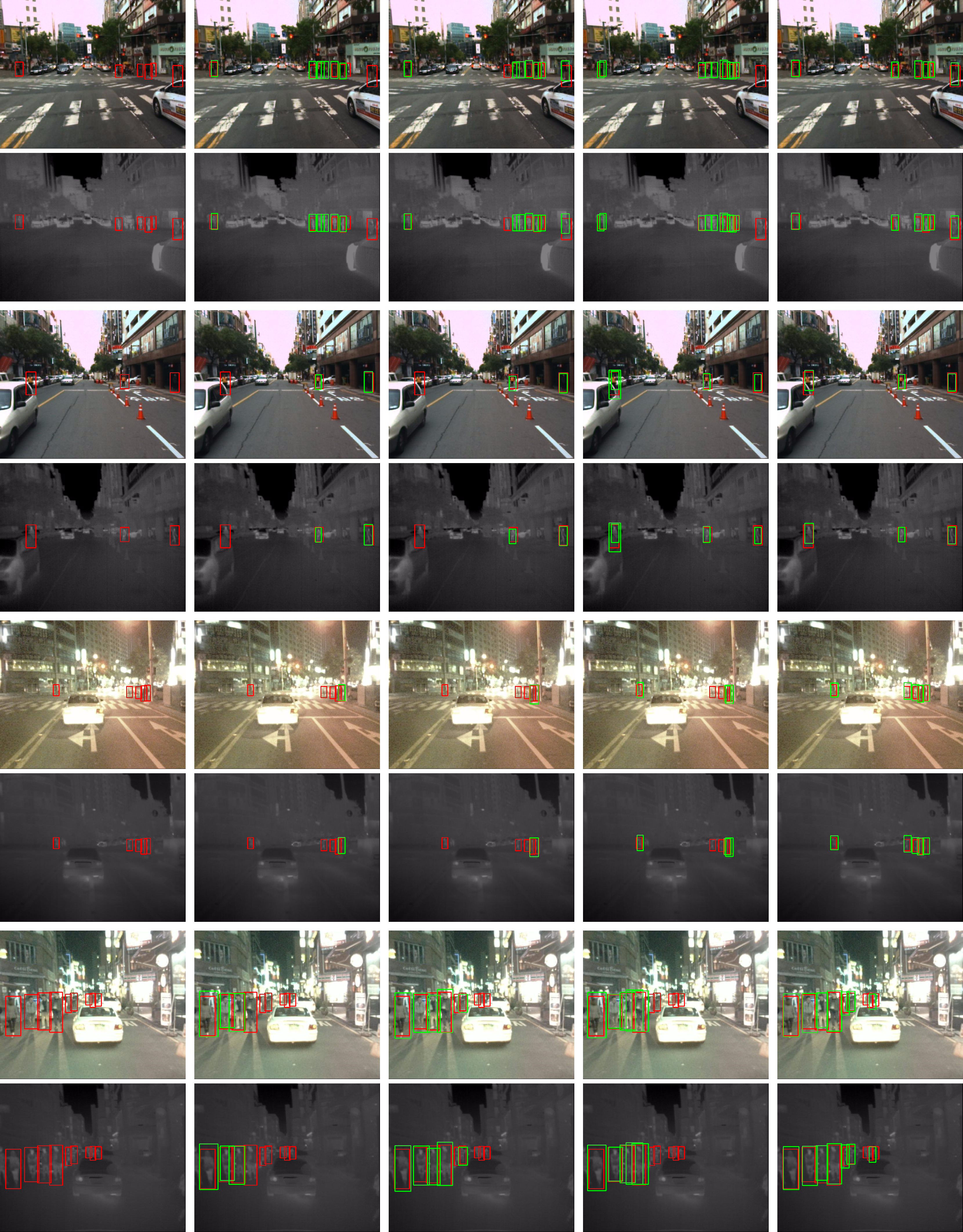}
\caption{Comparison of multispectral pedestrian detection results with other approaches. The first column shows the input pair images with ground-truth annotations depicted with red rectangles. The rest columns show the detection results (see green rectangles) of ACF+T+THOG \cite{hwang2015multispectral}, Halfway Fusion Faster R-CNN \cite{BMVC2016_73}, Fusion RPN + BF \cite{konig2017fully} and IAF R-CNN respectively. Our IAF R-CNN obtains a better overall detection accuracy than other three approaches.}
\label{fig:vis}       
\end{figure*}

To further demonstrate the effectiveness of the proposed IAF R-CNN, we illustrate several detection samples in Fig.~\ref{fig:vis}. Two samples are daytime images and the other two are nighttime images. The first column illustrates the input pair images and the rest three columns illustrate the detection results of ACF+T+THOG \cite{hwang2015multispectral}, Halfway Fusion Faster R-CNN \cite{BMVC2016_73}, Fusion RPN + BF \cite{konig2017fully} and IAF R-CNN. The red rectangles and the green rectangles depict the ground-truth bounding boxes and the predicted bounding boxes respectively. Detection results with FPPI 1 are presented. We can observe that IAF R-CNN can obtain superior detection results than ACF+T+THOG, Halfway Fusion Faster R-CNN and Fusion RPN+BF in some challenging cases under different illumination situations.

\section{Conclusion}
\label{conclude}
In this paper, we target the problem of multispectral pedestrian detection via convnets and make improvements in two aspects. First, we revisit several multispectral Faster R-CNN architectures and show that once properly adapted, these architectures can obtain promising improvements, some of which even achieve state-of-the-art performance on KAIST dataset. Second, we propose a novel Illumination-aware Faster R-CNN (IAF R-CNN) architecture which incorporates a color sub-network and a thermal sub-network into a unified framework by taking illumination conditions into consideration. An illumination-aware weighting mechanism is introduced to adaptively weight the detection confidence of two modalities according to the illumination measure and adaptively merge the two sub-networks to obtain final detections. Experimental results have demonstrated that the proposed IAF R-CNN is robust to different illumination conditions and outperforms all existing approaches on challenging KAIST dataset. 
In the future, we plan to further improve the robustness of our approach by fusing lidar data with multispectral images.

\section*{Acknowledgements}
The research is supported in part by NSFC (61572424) and the Science and Technology Department of Zhejiang Province (2018C01080). Min Tang is supported in part by NSFC (61572423,61732015) and Zhejiang Provincial NSFC (LZ16F020003).



\bibliographystyle{elsarticle-num} 
\bibliography{mybib}










\end{document}